\documentclass[10pt,twocolumn,letterpaper]{article}

\usepackage{iccv}
\usepackage{times}
\usepackage{epsfig}
\usepackage{graphicx}
\usepackage{amsmath}
\usepackage{amssymb}
\usepackage{bm}
\usepackage[percent]{overpic}
\usepackage{mathtools}
\usepackage{minibox}
\usepackage{subcaption}
\usepackage{hhline}
\usepackage{enumitem}
\usepackage{multirow}
\usepackage{floatpag}
\usepackage{makecell}
\usepackage{float}
\usepackage{afterpage}
\usepackage{booktabs}

\usepackage{array}
\newcolumntype{?}{!{\vrule width 1pt}}

\renewcommand{\etal}{\mbox{\textit{et al}.}}
\newcommand{\mref}[2]{\hyperref[#1]{\ref*{#1}#2}}

\ifx\supdraft\undefined
\newcommand{\supincludegraphics}[2][width=\linewidth]{\includegraphics[#1]{#2}}
\else
\newcommand{\supincludegraphics}[2][width=\linewidth]{\includegraphics[#1,draft]{#2}}
\fi

\usepackage[pagebackref=true,breaklinks=true,letterpaper=true,colorlinks,bookmarks=false]{hyperref}

\def\iccvPaperID{502} 

\iccvfinalcopy 

\ificcvfinal\fi

\usepackage[dvipsnames]{xcolor}

\begin{document}


\def\modelname{\mbox{PuppetGAN}}

\title{
\vspace*{-30px}
Cross-Domain Image Manipulation by Demonstration
}

\author{Ben Usman $^{1,2}$\\
{\tt\small usmn@bu.edu}
\and
Nick Dufour $^{1}$\\
{\tt\small ndufour@google.com} \\ \\
$^1$ Google AI
\and
Kate Saenko $^{2}$\\
{\tt\small saenko@bu.edu} \\ \\
$^2$ Boston University
\and
Chris Bregler $^{1}$\\
{\tt\small bregler@google.com}}

\maketitle

\setlength{\abovedisplayskip}{6pt}
\setlength{\belowdisplayskip}{4pt}

\begin{abstract}
In this work we propose a model that can manipulate individual visual attributes of objects in a real scene using examples of how respective attribute manipulations affect the output of a simulation. As an example, we train our model to manipulate the expression of a human face using nonphotorealistic 3D renders of a face with varied expression. Our model manages to preserve all other visual attributes of a real face, such as head orientation, even though this and other attributes are not labeled in either real or synthetic domain. Since our model learns to manipulate a specific property in isolation using only ``synthetic demonstrations'' of such manipulations without explicitly provided labels, it can be applied to shape, texture, lighting, and other properties that are difficult to measure or represent as real-valued vectors. We measure the degree to which our model preserves other attributes of a real image when a single specific attribute is manipulated. We use digit datasets to analyze how discrepancy in attribute distributions affects the performance of our model, and demonstrate results in a far more difficult setting: learning to manipulate real human faces using nonphotorealistic 3D renders.

\end{abstract}

\section{Introduction}

\begin{figure}[t]
\begin{center}
\hspace*{3px}\includegraphics[width=0.46\textwidth]{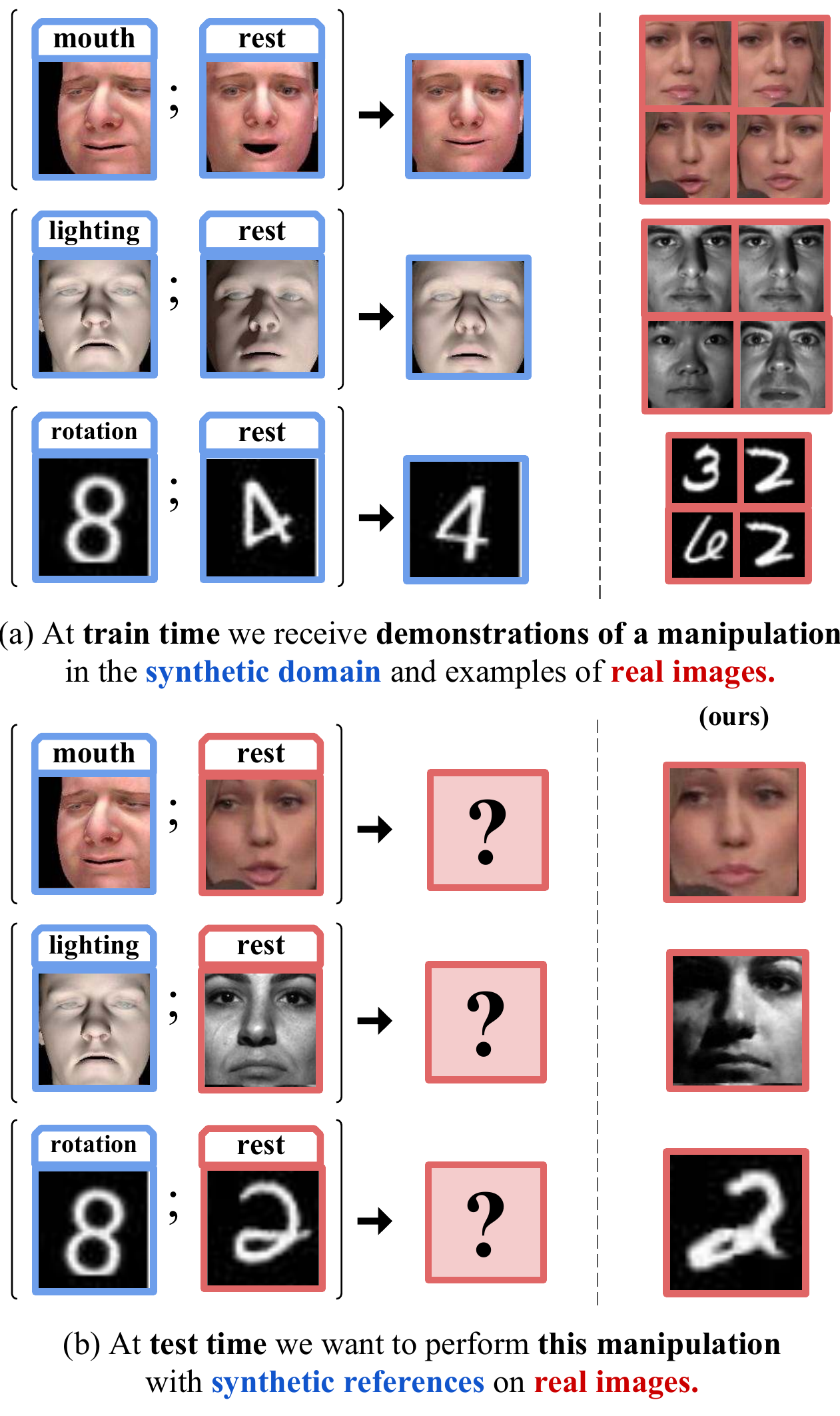}
\vspace*{-8px}
\caption{\textbf{Problem statement}: we want to manipulate real images using demonstrations of respective manipulations performed in a simulation, \eg to change mouth expression in real images using demonstrations of mouth expression manipulation in nonphotorealistic face 3D renders, or to relight real faces from examples of relighted synthetic faces, or to rotate hand-written digits using rotated typewritten digits. The proposed \modelname\ model correctly manipulates real images and uses only unlabeled real images and synthetic demonstrations during training. Video: {\small \url{http://bit.ly/arxiv_puppet}}. \label{fig:problem}
}
\vspace*{-40px}
\end{center}
\end{figure}

\begin{figure*}
\thisfloatpagestyle{empty}
\begin{center}
\vspace{-25px}
\includegraphics[width=0.9\textwidth,trim={0 130 200 0},clip]{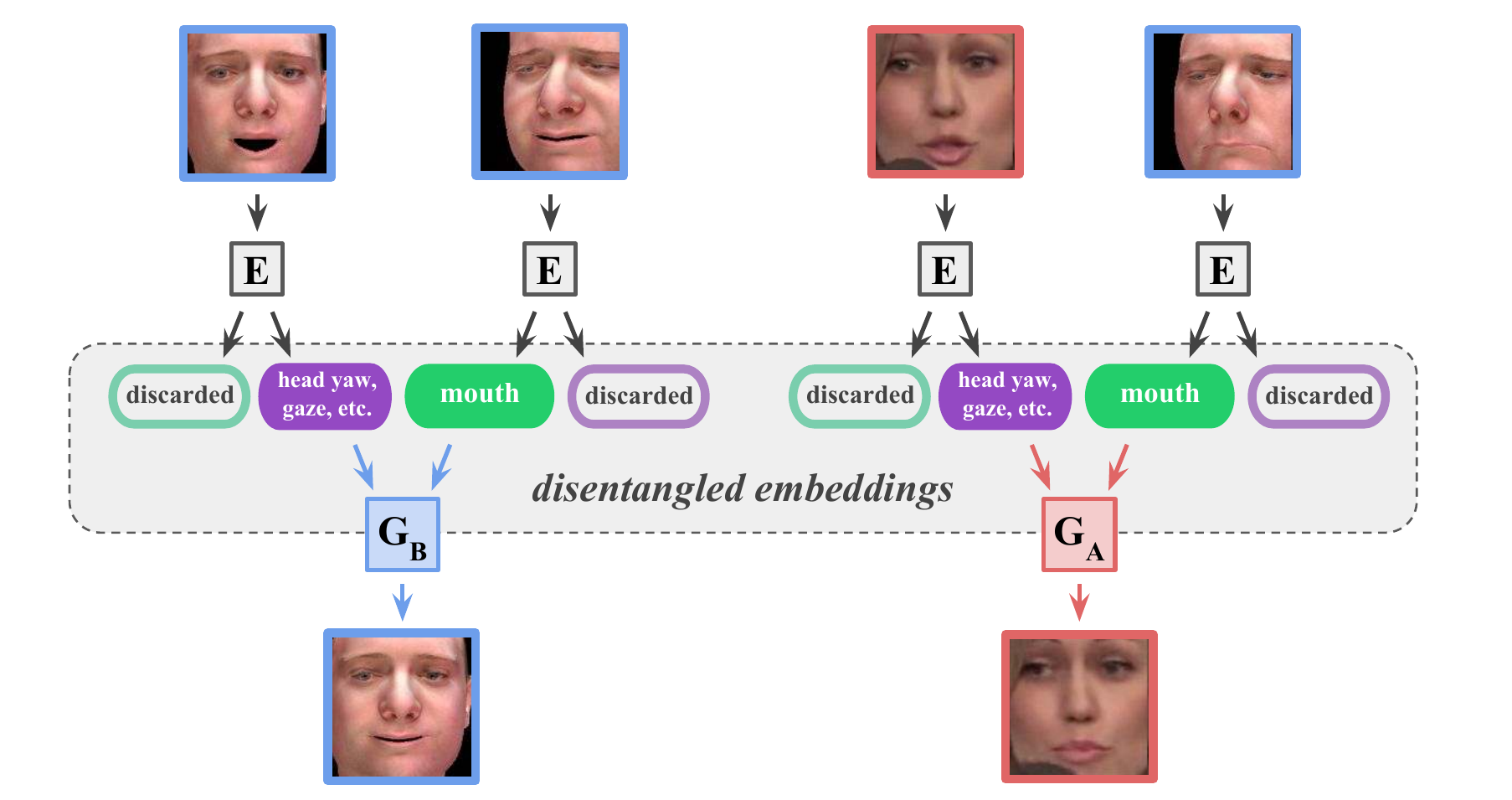}\vspace*{-10px}
\caption{\textbf{PuppetGAN overview:} we train a domain-agnostic encoder ($E$), a decoder for the real domain ($G_A$) and a decoder for the synthetic domain ($G_B$) to disentangle the the attribute we would like to control in real images (the ``attribute of interest'' - AoI - mouth expression in this example), and all other attributes (head orientation, gaze direction, microphone position in this example) that are not labeled or even not present (\eg microphone) in the synthetic domain. Our model is trained on demonstrations of how the AoI is manipulated in synthetic images and individual examples of real images. At \textit{test time}, a real image can be manipulated with a synthetic reference input by applying a real decoder to the attribute embedding of the reference image ({\color{ForestGreen} green} capsule) combined with the remaining embedding part ({\color{Plum}purple} capsule) of the real input. \label{fig:simple}}

\vspace{5px}

\hspace*{-15px}\includegraphics[width=1.05\textwidth,trim={15 0 20 0},clip]{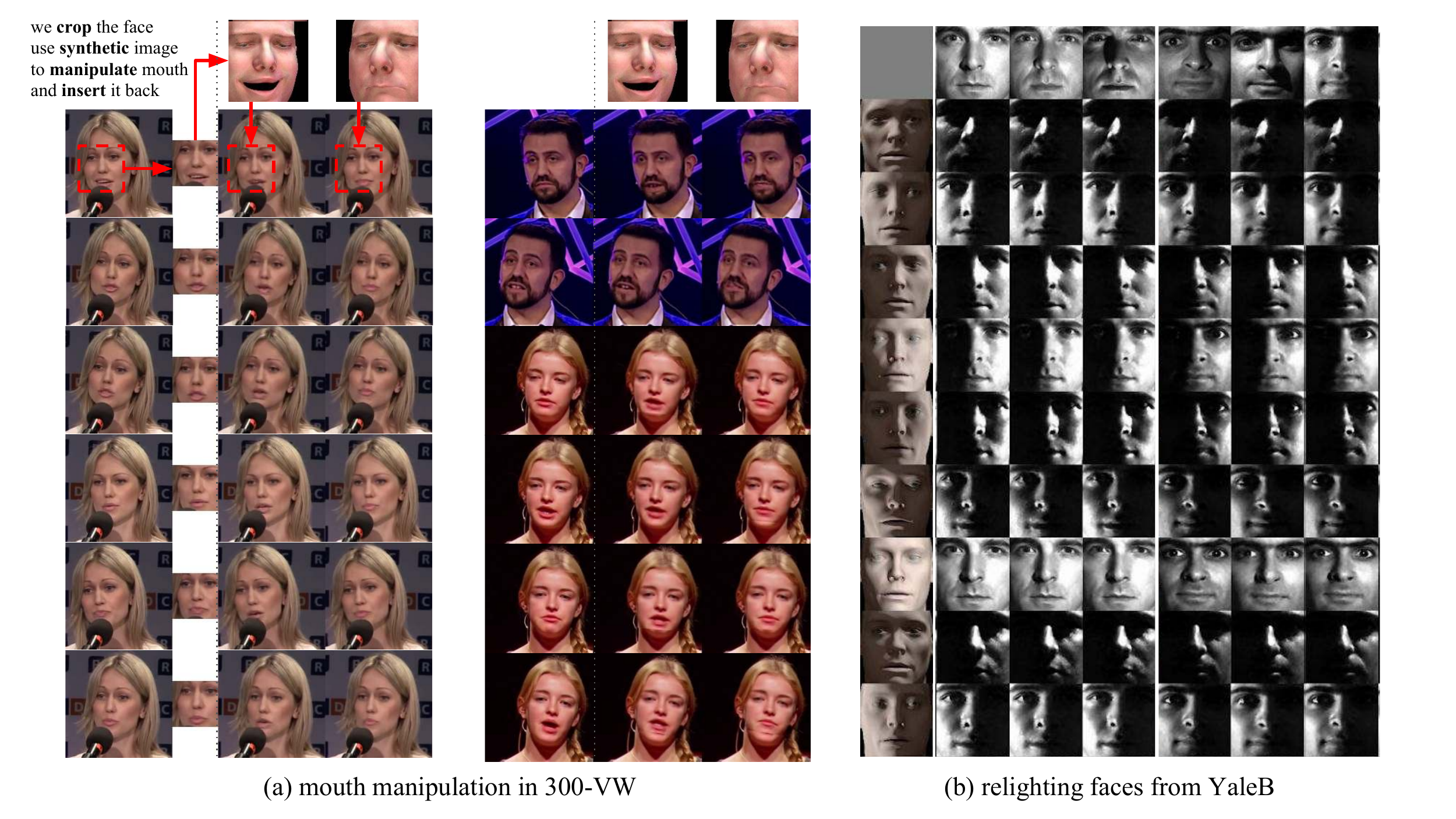}
\vspace{-20px}\caption{ More examples with other identities are provided in the supplementary. (a) When trained on face crops from a single 300-VW \cite{sagonas2016300} video, PuppetGAN learns to manipulate mouth expression while preserving head orientation, gaze orientation, expression, \textit{etc.} so well that directly ``pasting'' the manipulated image crop back into the frame without any stitching yields realistically manipulated images without noticeable head orientation or lighting artifacts (chin stitching artifacts area are unavoidable unless an external stitching algorithm is used); the video demonstration is available in the supplementary and at {\small \url{http://bit.ly/arxiv_puppet}}.
(b) When trained on face crops of all subjects from YaleB \cite{georghiades2001few} combined into a single domain, PuppetGAN learns to properly apply lighting (AoI) from a synthetic reference image and correctly preserves subjects' identities without any identity labels; lighting of the original real image has little to no effect on the output. 
\label{fig:faces}}

\end{center}

\label{fig:short}
\end{figure*}

Recent unsupervised image-to-image translation models \cite{liu2017unsupervised,royer2017xgan,CycleGAN2017} demonstrated an outstanding ability to learn semantic correspondences between images from visually distinct domains. Using these models is especially fruitful for domains that lack large labeled datasets since converting an output of an existing simulation to an image that closely resembles the real domain gives rise to a virtually infinite source of labeled training data.  Unfortunately, since these models receive no supervision relating semantic structure of two domains, latent encodings of visual attributes are strongly entangled and can not be manipulated independently. For example, CycleGAN \cite{CycleGAN2017} can not be easily used to extract mouth expression from a synthetically rendered face image and ``apply'' it to an image of a real person since all attributes (face orientation, expression, lighting) are entangled. The lack of such control limits usage of image-to-image translation methods in many areas where manipulation of individual attributes is required, but vast labeled dataset can not be collected for each possible input domain, including augmented reality, virtual avatars, or semantic video post-processing.



Several methods have been proposed to enable users to exert attribute-level control over the generated data, but these methods either require a precise model (i.e. precise simulations) of the target domain \cite{Thies2016} or detailed attribute labels and suitable means of training a label estimator for both domains \cite{Cao2018dida,liu2018detach}. Building such estimators is not an easy task when the desired attribute(s) are difficult to measure or even represent numerically as inputs to a neural model, such as: global illumination, texture, or shape. Unsupervised cross-domain disentanglement methods \cite{huang2018munit,DRIT} on the other hand, do not provide means for specifying which of the shared ``content'' attributes should be altered. 

To overcome these limitations, we propose ``\modelname,'' a deep model for targeted and controlled modification of natural images that requires neither explicit attribute labels nor a precise simulation of the real domain. To enable control over a specific attribute in real images \modelname\ only requires examples (``synthetic demonstrations'') of how the desired attribute manipulation affects the output of a crude simulation. It uses these synthetic demonstrations to supervise attribute disentanglement in the synthetic domain, and extends this disentanglement to the real domain by specifying which attributes are supposed to be preserved by multiple consecutive attribute manipulations. We quantitatively evaluate how well our model can preserve other attributes of the input when a single attribute is manipulated. In this work we:

\begin{enumerate}
    \item Introduce a new challenging ``cross-domain image manipulation by demonstration'' task: manipulate a specific attribute of a real image via a synthetic reference image using \textit{only} examples of real images and demonstrations of the desired attribute manipulation in \textit{a synthetic domain} at train time in the presence of a significant domain shift both in the domain appearance and attribute distributions.
    \item Propose a model that enables controlled manipulation of a specific attribute and correctly preserves other attributes of the real input. We are first to propose a model that enables this level of control under such data constraints at train time.
    \item Propose both proof-of-concept (digits) and realistic (faces and face renders) dataset pairs [to be published online] and a set of metrics for this task. We are first to quantitatively evaluate effects of cross-domain disentanglement on values of other (non-manipulated) attributes of images. 
\end{enumerate}

\section{Related work}

\textbf{Model-based Approaches}. Thanks to recent advances in differentiable graphics pipelines \cite{loper2014diffgraph}, generative neural models \cite{goodfellow2014generative}, and high-quality morphable models \cite{paysan20093d}, great strides have been made in controlled neural image manipulation. For example, the work of Thies \textit{et al.} \cite{Thies2016} proposed a way to perform photo-realistic face expression manipulation and reenactment that cannot be reliably detected even by trained individuals. Unfortunately, methods like these rely on precise parametric models of the target domain and accurate fitting of these parametric models to input data: in order to manipulate a single property of an input, all other properties (such as head pose, lighting and face expression in case of face manipulation) have to be estimated from an image and passed to a generative model together with a modified attribute to essentially ``rerender'' a new image from scratch. This approach enables visually superb image manipulation, but requires a detailed domain model capable of precisely modeling all aspects of the domain and re-rendering any input image from a vector of its attributes - it is a challenging task, and its solution often does not generalize to other domains. 
Our model merely requires a crude domain model, does not require any parameter estimation from an input image, and therefore can be applied in many more contexts, achieving a good trade-off between image fidelity and costs associated with building a highly specialized domain model. 

\begin{table}[t]
\centering
\small
\begin{tabular}{ccc}
\toprule
    \textbf{\makecell{attribute \\ labels for}} & \textbf{single-domain} & \textbf{cross-domain} \\ \midrule
\makecell{single domain} & \makecell{Mathieu~\etal~\cite{mathieu2016disentangling}, \\ Cycle-VAE \cite{harsh2018disentangling}, \\ Szab\'o~\etal~\cite{szabo2017challenges}}& \makecell{E-CDRD \cite{liu2018detach}, \\ DiDA \cite{Cao2018dida}, \\ \textbf{PuppetGAN}} \\ \addlinespace
\makecell{both domains} & --- & UFDN \cite{Liu2018ufdn} \\ \addlinespace
unsupervised & \makecell{InfoGAN \cite{chen2016infogan}, \\ $\beta$-VAE \cite{higgins2017betavae}, \\ $\beta$-TCVAE \cite{chen2018tcvae}} & \makecell{DRIT \cite{DRIT}, \\ MUNIT \cite{huang2018munit}} \\ \bottomrule
\end{tabular}
\caption{Some of existing disentanglement methods that enable controlled manipulation of real images.\vspace{-10px}}
\label{tab:my_label}


\end{table}

\textbf{Single-Domain Disentanglement}. One alternative to full domain simulation is learning a representation of the domain in which the property of interest and other properties could be manipulated independently - a so called ``disentangled representation''. We summarized several kinds of disentanglement methods that enable such control over real images using simulated examples in Table 1. Supervised single-domain disentanglement methods require either explicit or weak (pairwise similarity) labels \cite{harsh2018disentangling,mathieu2016disentangling,szabo2017challenges} for real images - a much stronger data requirement then the one we consider. As discussed in the seminal work of Mathieu~\etal~\cite{mathieu2016disentangling} on disentangling representations using adversarial learning and partial attribute labels and later explored in more details by Szab\'o~\etal~\cite{szabo2017challenges} and Harsh Jha~\etal~ \cite{harsh2018disentangling}, there are always degenerate solutions that satisfy proposed constraints, but cheat by ignoring one component of the embedding and hiding information in the other, we discuss steps we undertook to combat these solutions in the model and experiment sections.

Unsupervised single-domain methods \cite{chen2018tcvae,chen2016infogan,higgins2017betavae} enable visually impressive manipulation of image properties without any attribute labels, but do not provide the means to select a \textit{specific} property that we want to control \textit{a priori} - the end user is left to the mercy of the model that might not necessarily disentangle the specific property he or she is interested in.

\textbf{Unsupervised Cross-Domain Disentanglement}. Recently proposed unsupervised cross-domain disentanglement methods \cite{huang2018munit,DRIT} focus on disentangling domain specific properties (often corresponding to the ``style'' of the domain) from those shared by both domains (``content''), therefore providing tools for manipulating appearance of a particular image while preserving the underlying structure in a completely unsupervised fashion. Our approach, however, can disentangle and independently manipulate a single specific ``content'' attribute (\eg face expression) even if other ``content'' attributes (\eg head orientation, lighting) significantly vary in both real and synthetic domains, therefore enabling much finer control over the resulting image.

\textbf{Supervised Cross-Domain Disentanglement}. In the presence of the second domain, one intuitive way of addressing the visual discrepancy between the two is to treat the domain label as just another attribute \cite{Liu2018ufdn} and perform disentanglement on the resulting single large partially labeled domain. This approach enables interpolation between domains, and training conditional generative models using labels from a single domain, but does not provide means for manipulation of existing images across domains, unless explicit labels in both domains are provided. Recent papers \cite{Cao2018dida,liu2018detach} suggested using explicit categorical labels to train an explicit attribute classifiers on synthetic domain and adapt it to the real domain; the resulting classifier is used to (either jointly or in stages) disentangle embeddings of real images. These works showed promising results in manipulating categorical attributes of images to augment existing dataset (like face attributes in CelebA \cite{liu2015deep} or class label in MNIST), but neither of these methods was specifically designed for or tested for their ability to \textit{preserve other attributes of an image}: if we disentangle the size of a digit from its class for the purpose of, effectively, generating more target training samples for classification, we do not care whether the size is preserved when we manipulate the digit class, since that would still yield a correctly ``pseudo-labeled'' sample from the real domain. Therefore, high classification accuracies of adapted attribute classifiers (reported in these papers) do not guarantee the quality of disentanglement and the ability of these models to preserve unlabeled attributes of the input. Moreover, these method requires explicit labels making them not applicable to wide range of attributes that are hard to express as categorical labels (shape, texture, lighting). In this work, we specifically focus on manipulating individual attributes of images using demonstrations from another domain, in the presence of a significant domain shift (both visual and in terms of distributions of attribute values) and explicitly quantitatively evaluate the ability of our model to preserve all attributes other the the one we manipulated. 




\section{Method}

\begin{figure*}[ht]
\centering
\vspace*{-5px}
\begin{overpic}[width=\textwidth]{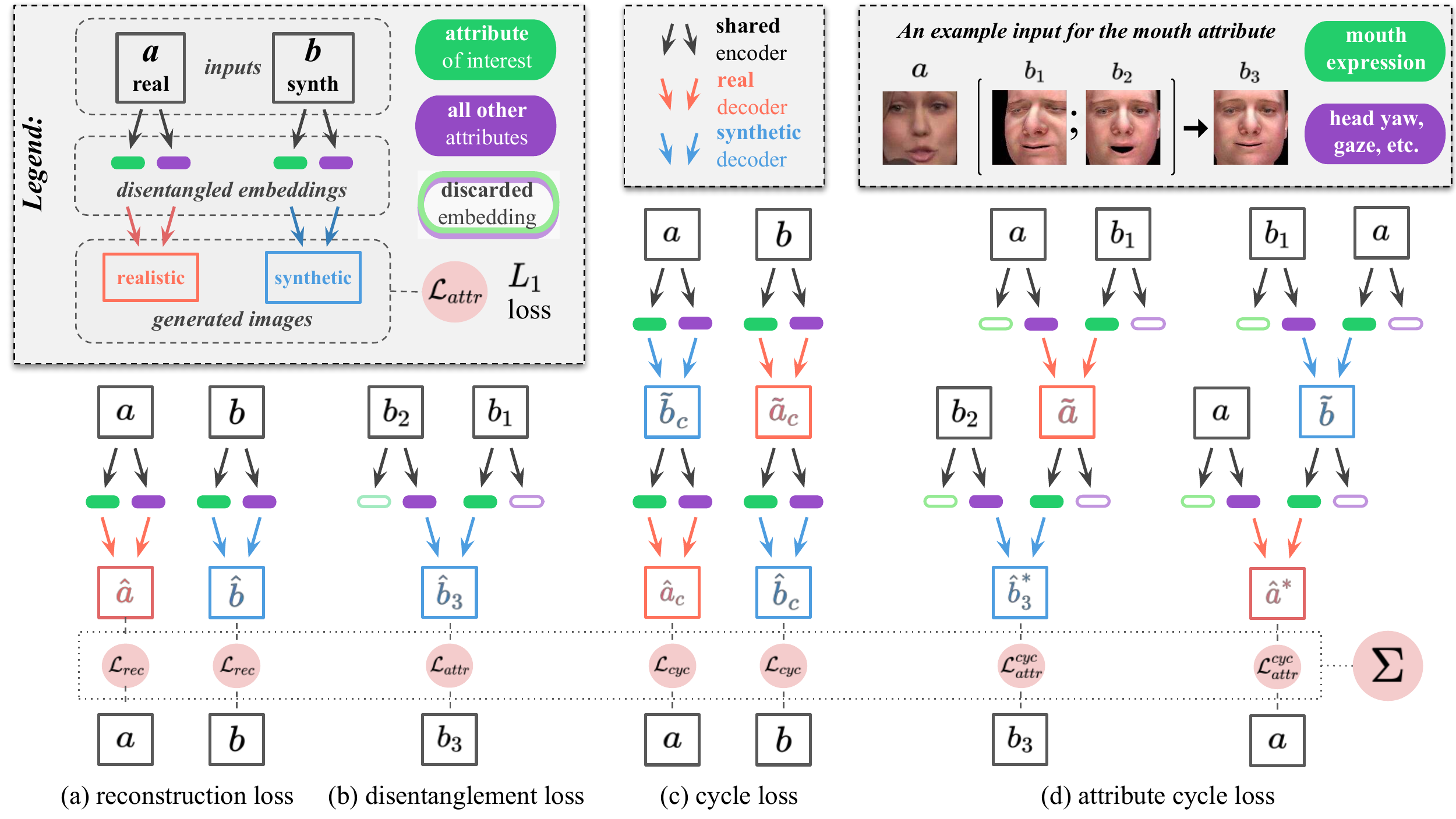}
\end{overpic}

\caption{\textit{(Best viewed in color, a color blind friendly version is available in the supplementary).} Supervised losses jointly optimized during training of the PuppetGAN. When combined, these losses ensure that the ``attribute embedding'' ({\color{ForestGreen} green} capsule) affects only the attribute of interest (AoI) in generated images, and that the ``rest embedding'' ({\color{Plum}purple} capsule) does not affect the AoI in generated images. When trained, manipulation of AoI in real images can be performed by replacing their attribute embedding components. Unsupervised (GAN) losses are not shown in this picture. An example at the top right corner illustrates sample images fed into the network to disentangle mouth expression (AoI) from other face attributes in real faces. Section 3 provides more details on the intuition behind of these losses. \vspace*{-10px}}
\label{fig:model}
\end{figure*}

In this section we formally introduce our data constraints, define a disentangled encoder and domain decoders used in the loss, and describe a set of constraints that ensure proper disentanglement of synthetic images and extension of this disentanglement to a real domain. We find ``must be equal'' notation more concise for our purposes, i.e. $y = f(x) \ \ \forall (x, y) \in D$ in constraint definitions below corresponds to the $\sum_{(x, y) \in D}|| y - f(x) ||$ penalty in the loss. 

\textbf{Setup.} Consider having access to individual real images $a \in \mathcal X_A$, and triplets of synthetic images $(b_1, b_2, b_3) \in \mathcal X_B$ such that $(b_1, b_3)$ share the attribute of interest (AoI - the attribute that we want to control in real images), whereas the pair $(b_2, b_3)$ shares all other attributes present in the synthetic domain. See top right corner of Figure \ref{fig:model} for an example of inputs fed into the network to learn to control mouth expression (AoI) in real faces using crude face renders. 

\textbf{Model.} The learned image representation consists of two real-valued vectors $e_{attr}$ and $e_{rest}$ denoted as green and purple capsules in Figures \ref{fig:simple} and \ref{fig:model}. We introduce domain-agnostic encoders for the attribute of interest $E_{attr}$ and all other attributes $E_{rest}$, and two domain-specific decoders $G_A, G_B$ for the real and synthetic domains respectively:
\begin{gather*}
    E_{attr}: (x) \mapsto e_{attr},  \quad  G_A: (e_{attr}, e_{rest}) \mapsto x_a\\
    E_{rest}: (x) \mapsto e_{rest}, \quad G_B: (e_{attr}, e_{rest})  \mapsto x_b.
\end{gather*}

To simplify the loss definitions below, we introduce the domain-specific ``attribute combination operator'' that takes a pair of images $(x, y)$, each from either of two domains, combines embeddings of these images, and decodes them as an image in the specified domain $K$:
\begin{equation*}
    C_K(x, y) \triangleq G_K\big(E_{attr}(x), E_{rest}(y)\big), \ K \in \{A, B\}.
\end{equation*}

\textbf{Losses.} We would like $C_K(x, y)$ to have the AoI of $x$ and all other attributes of $y$, but we can not enforce this directly as we did not introduce any explicit labels. Instead we \textit{jointly} minimize the weighted sum of $L_1$-penalties for violating the following constraints illustrated in Figure \ref{fig:model} with respect to all parameters of both encoders and decoders:
\begin{enumerate}[label=(\alph*),itemsep=-0.5ex]
\item the reconstruction constraint ensures that encoder-decoder pairs actually learn representations of respective domains
$$x = C_K(x, x) \quad \forall x \in \mathcal{X}_K, \ \forall K \in \{A, B\}$$
\item the disentanglement constraint ensures correct disentanglement of synthetic images by the shared encoder and the decoder for the synthetic domain
$$ b_3 = C_B(b_1, b_2) \quad \forall (b_1, b_2, b_3) \in \mathcal{X}_B$$
\item the cycle constraint was shown \cite{CycleGAN2017} to improve semantic consistency in visual correspondences learned by unsupervised image-to-image translation models
\begin{gather*}
    a = C_A(\tilde{b}_c, \tilde{b}_c), \ \ \text{ where } \ \ \tilde{b}_c = C_B(a, a)  \\
    b = C_B(\tilde{a}_c, \tilde{a}_c), \ \ \text{ where } \ \ \tilde{a}_c = C_A(b, b) \\
    \forall a \in \mathcal{X}_A \quad \forall b \in \mathcal{X}_B
\end{gather*}
\item the pair of attribute cycle constraints prevents shared encoders and the real decoder $G_A$ from converging to a degenerate solution - decoding the entire real image from a single embedding and completely ignoring the other part. The first ``attribute cycle constraint'' (the left column in Figure~\mref{fig:model}{d}) ensures that the first argument of $C_A$ is not discarded:
\begin{gather*}
b_3 = C_B(\tilde{a}, b_2), \ \ \text{ where } \ \ \tilde{a} = C_A(b_1, a) \\
\forall a \in \mathcal{X}_A, \ \forall (b_1, b_2, b_3) \in \mathcal{X}_B.
\end{gather*}
The only thing that is important about $\tilde a$ as the first argument of $C_B$ is its attribute value, so $C_A$ must not discard the attribute value of its \textit{first} argument $b_1$, since otherwise reconstruction of $b_3$ would become impossible. The ``rest'' component of $a$ should not influence the estimate of $b_3$ since it only affects the ``rest'' component of $\tilde{a}$ that is discarded by later application of $C_B$. To ensure that the second ``rest embedding'' argument of $C_A$ is not always discarded, the second attribute cycle constraint (the right column in Figure~\mref{fig:model}{d})
\begin{gather*}
a = C_A(\tilde{b}, a), \ \ \text{ where } \ \ \tilde{b} = C_B(a, b) \\ 
\forall a \in \mathcal{X}_A, \ b \in \mathcal{X}_B
\end{gather*}
penalizes $C_A$ if it ignores its \textit{second} argument since the ``rest'' of $a$ is not recorded in $\tilde{b}$ and therefore can be obtained by $C_A$ only from its second argument.
\end{enumerate}

The proposed method can be easily extended to disentangle multiple attributes at once using separate encoders and example triplets for each attribute. For example, to disentangle two attributes $p$ and $q$ using encoders $E^p_{attr}, E^q_{attr}$ and synthetic triplets $(b^p_1, b^p_2, b^p_3), (b^q_1, b^q_2, b^q_3)$ where $(b^p_2, b^p_3)$ share all other attributes except $p$ (\textit{including} $q$), and vice versa, the disentanglement constraint should look like: 
\begin{gather*}
    b^p_3 = G_B(E^p_{attr}(b^p_1), E^q_{attr}(b^p_2), E_{rest}(b^p_2)) \\
    b^q_3 = G_B(E^p_{attr}(b^q_2), E^q_{attr}(b^q_1), E_{rest}(b^q_2)).
\end{gather*}

In addition to supervised losses described above we apply unsupervised adversarial LS-GAN \cite{mao2017least} losses to all generated images. Discriminators $D_K(x')$ and attribute combination operators $C_K(x, y)$ are trained in an adversarial fashion so that any combination of embeddings extracted from images $x, y$ from either of two domains and decoded via either real or synthetic decoder $G_K$ looks like a reasonable sample from the respective domain. Other technical implementation details are provided in the supplementary. 
\begin{figure*}
\thisfloatpagestyle{empty}
\begin{center}
\vspace*{-20px}\hspace*{-15px}\includegraphics[width=1\textwidth,trim={0 30 0 13},clip]{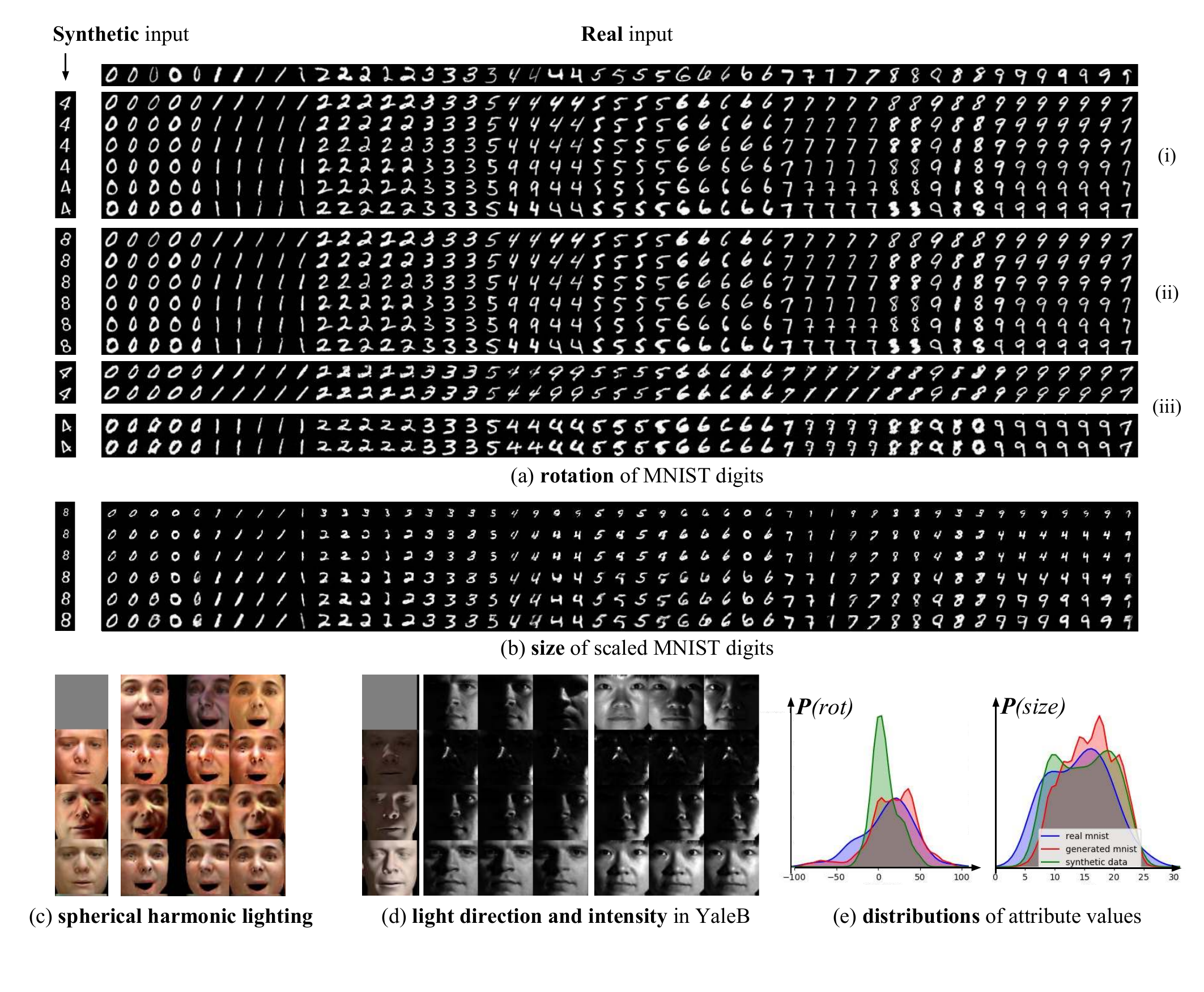}
\vspace*{-3px}\caption{\textbf{(a-d)} The PuppetGAN model correctly manipulates a single specified attribute the real input and preserves other attributes of that input without any attribute labels at train time. \textbf{(ii)} Moreover, other attributes of the ``synthetic reference input'' have no effect on the output. \textbf{(iii)} The proposed model ``saturates'' when the AoI in the synthetic input is beyond limits observed during training. \textbf{(e)} The distribution of attributes is monotonically remapped to match the real domain. \label{fig:results}}
\includegraphics[width=0.95\textwidth,trim={0 70 0 0},clip]{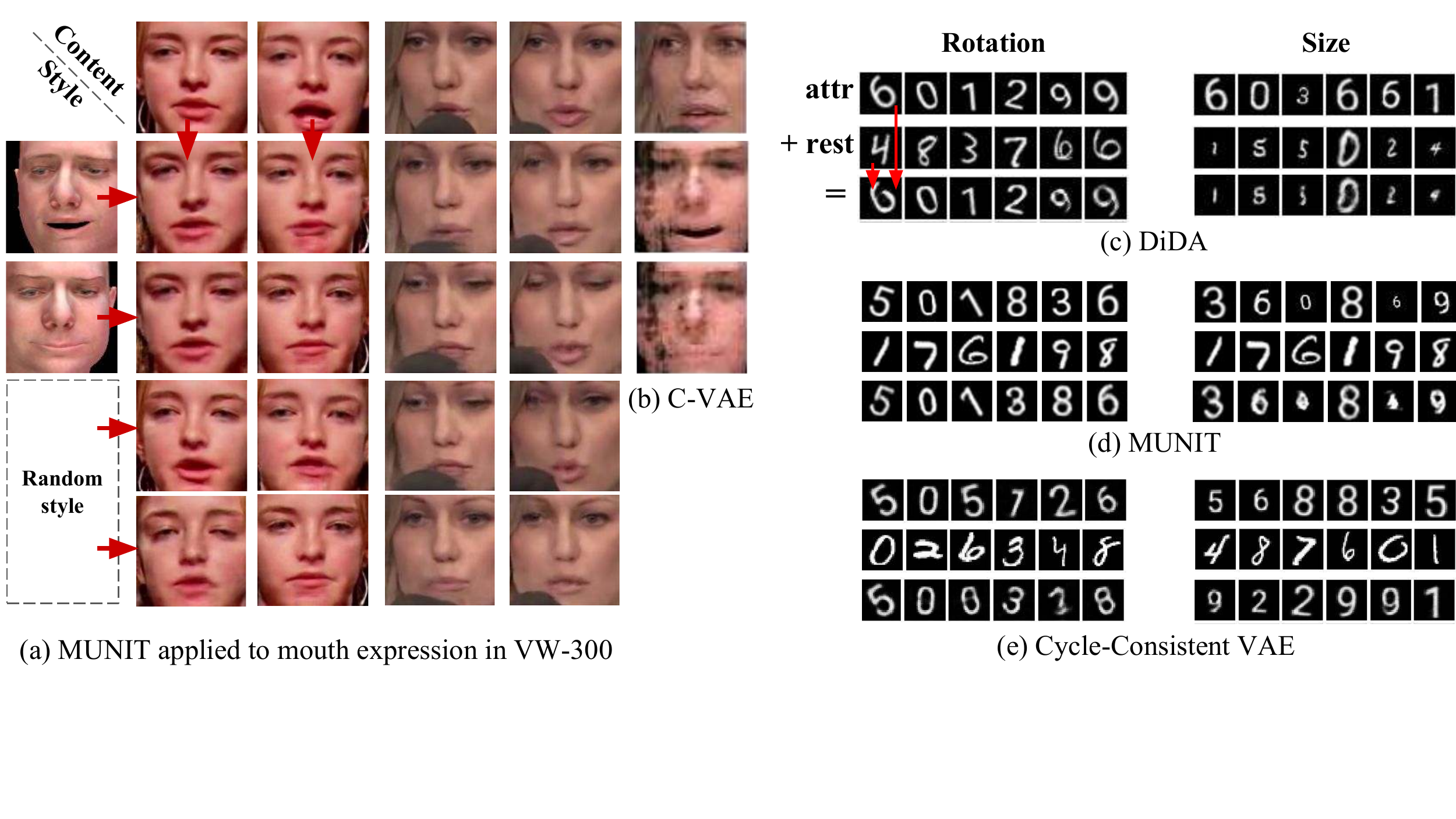}
\vspace*{-8px}\caption{\textbf{Related methods} (only DiDA is directly comparable) \textbf{(a)} MUNIT disentangled mouth expression from head orientation, but style spaces of two domains are not aligned, so controlled mouth manipulation is not possible; \textbf{(b)} Cycle-Consistent VAE is not suited for large domains shift; \textbf{(c)} DiDA converged to degenerate solutions that used only one input; \textbf{(d)} MUNIT disentangled stroke from other attributes (\ie did not isolate rotation or size from the class label); \textbf{(e)} Cycle-Consistent VAE was able to extract information only from real inputs that looked ``synthetic enougth''. \label{fig:munit}}
\end{center}
\end{figure*}

\section{Experiments and Results}

\textbf{Setup.} We evaluated the ability of our model to disentangle and manipulate individual attributes of real images using synthetic demonstrations in multiple different settings illustrated in Figures \ref{fig:faces} and \ref{fig:results}.

\begin{table*}
\vspace*{-15px}
\footnotesize
\centering


\begin{tabular}{lcccccccccc}
\toprule
\multicolumn{5}{c}{\bfseries Training Conditions}  & \multicolumn{6}{|c}{\bfseries Results} \\
 \cmidrule(lr){1-5} \cmidrule(lr){6-11}
Model & Modified Data & Attribute & $J_\text{attr}^{\text{syn}}$ & $J_\text{rest}^{\text{syn}}$ & $J_\text{attr}^{\text{gen}}$ & $J_\text{rest}^{\text{gen}}$ & Acc $\uparrow$ & $r^{\text{syn}}_{\text{attr}} \uparrow$ & $ r^{\text{syn}}_{\text{rest}} \downarrow $ & $V_{\text{rest}} \downarrow$ \\
\midrule
PuppetGAN & & \multirow{2}{*}{Rot} & \multirow{2}{*}{0.05} & \multirow{2}{*}{2.2} & 0.03 & 0.32 & 0.97 & 0.40 & 0.11 & 0.01 \\
CycleGAN & & & && 0.05 & 0.28 & 0.11 & 0.54 & 0.37 & 0.33 \\
PuppetGAN &  & \multirow{2}{*}{Size} & \multirow{2}{*}{0.27} & \multirow{2}{*}{0.78} & 0.24 & 0.04 & 0.73 & 0.85 & 0.02 & 0.02 \\
CycleGAN & &  & & & 0.20 & 0.07 & 0.10 & 0.28 & 0.06 & 0.28 \\
\addlinespace
PuppetGAN & smaller synth & Rot & 0.06 & 108 & $+\infty$ & $+\infty$ & 0.10 & 0.06 & 0.04 & 0.01 \\
PuppetGAN & unscaled real & Size & 0.90 & 0.92 & 0.27 & 0.05 & 0.64 & 0.28 & 0.07 & 0.01 \\
\bottomrule
\end{tabular}

\vspace{-3px}\caption{\small The quality of attribute manipulation and isolation for rotation and scaling of MNIST digits (Figure \mref{fig:results}{ab}). The PuppetGAN performs significantly better then an analogous non-disentangled baseline (CycleGAN). We measure how well models preserve the class labels of real inputs (Acc), AoI of synthetic inputs $r^{\text{synth}}_{\text{attr}}$, and ignore non-AoI of synthetic inputs $r^{\text{synth}}_{\text{rest}}$. We investigate how larger discrepancy between synthetic and real attribute distributions $J^{\text{syn}}$ affects the performance. Arrows $\uparrow\downarrow$ indicate if higher or lower values are better. \label{tab:results}}
\vspace*{-20px}
\end{table*}

\begin{enumerate} [wide, labelwidth=!, labelindent=0pt,itemsep=-0.5ex,topsep=0pt]
    \item \textit{Size and rotation of real digits} from MNIST and USPS were manipulated using a synthetic dataset of typewritten digits rendered using a sans-serif Roboto font. 
    \item \textit{Mouth expression in human face crops} from the \mbox{VW-300} \cite{sagonas2016300} dataset was manipulated using synthetic face renders with varying face orientation and expression, but same identity and lighting, obtained using Basel parametric face model \cite{kortylewski2018training,kortylewski2018empirically} with the global illumination prior \cite{egger2018occlusion}.
    \item \textit{Global illumination} (spherical harmonics) in female synthetic face renders was manipulated using male renders with different head orientation and expression.
    \item \textit{Direction and power of the light source} in real faces from the YaleB \cite{georghiades2001few} dataset were manipulated using synthetic 3D face renders with varying lighting and identities (but constant expression and head orientation).
\end{enumerate}
We used visually similar digit dataset pairs to investigate how discrepancy in attribute distributions affects the performance of the model, \eg how it would perform if synthetic digits looked similar to real digits, but were much smaller then real ones or rotated differently. In face manipulation experiments we used a much more visually distinct synthetic domain. In VW-300 experiments we treated each identity as a separate domain, so the model had to learn to preserve head orientation and expression of the real input; we used the same set of 3D face renders across all real identities. In the experiment on reapplying environmental lighting to synthetic faces, expression and head orientation of the input had to be preserved. In the lighting manipulation experiment on the YaleB dataset we used a single large real domain with face crops of many individuals with different lighting setups each having the same face orientation across the dataset, so the model had to learn to disentangle and preserve the identity of the real input.

\textbf{Metrics.} In order to quantitatively evaluate the performance of our model on digits we evaluated Pearson correlation ($r$) between measured attribute values in inputs and generated images. We measured rotation and size of both input and generated digit images using image moments, and trained a LeNet \cite{lecun1998gradient} to predict digit class attribute. Below we define metrics reported in the Table \ref{tab:results}. The AoI measurements in images generated by an ``ideal'' model should strongly correlate with the AoI measurements in respective synthetic inputs ($r^{\text{syn}}_{\text{attr}} \uparrow$ - the arrow direction indicates if larger or smaller values of this metric is ``better''), and the measurement of other attributes should strongly correlate with those in real inputs (Acc - accuracy of preserving the digit class label - higher is better), and no other correlations should be present ($r^{\text{syn}}_{\text{rest}}$ lower is better). For example, in digit rotation experiments we would like the rotation of the generated digit to be strongly correlated with the rotation of the synthetic input and uncorrelated with other attributes of the synthetic input (size, class label, etc.); we want the opposite for real inputs. Also, if we use a different synthetic input with the same AoI value (and random non-AoI values) there should be no change in pixel intensities in the generated output (small variance $V_{\text{rest}}$). Optimal values of these metrics are often unachievable in practise since attributes of real images are not independent, \eg inclination of real digits is naturally coupled with their class label (sevens are more inclined then twos), so preserving the class label of the real input inevitably leads to non-zero correlation between rotation measurements in real and generated images. We also estimated discrepancy in attribute distributions by computing Jensen-Shannon divergence between optimal \cite{scott2012multivariate} kernel density estimators of respective attribute measurements between real and synthetic images ($J^{\text{syn}}$) as well as real and generated images ($J^{\text{gen}})$. In order to quantitatively evaluate to what extent proposed disentanglement losses improve the quality of attribute manipulation, we report same metrics for an analogous model without disentanglement losses that translates all attributes of the synthetic input to the real domain (CycleGAN).

\textbf{Hyperparameters.} We did not change any hyperparameters across tasks, the model performed well with the initial ``reasonable'' choice of parameters listed in the supplementary. As all adversarial methods, our model is sensitive to the choice of generator and discriminator learning rates.

\textbf{Results.} The proposed model successfully learned to disentangle the attribute of interest (AoI) and enabled isolated manipulation of this attribute using embeddings of synthetic images in all considered experiment settings:
\begin{enumerate}[wide, labelwidth=!, labelindent=0pt,itemsep=-0.5ex,topsep=0pt]
    \item In the digit rotation experiment (Figure \mref{fig:results}{a}), generated images had the class label, size and style of the respective real input and rotation of the respective synthetic input, and did not change if either class or size of the synthetic (Figure \mref{fig:results}{(ii)}), or rotation of the real input changed. Quantitative results in Table \ref{tab:results} indicate that proposed losses greatly improve the quality of isolated attribute manipulation over a non-disentangled baseline. Attributes were properly disentangled in all face manipulation experiments (Figure \mref{fig:faces}{ab}, \mref{fig:results}{cd}), \eg in the YaleB experiment ``original'' lighting of real faces and identities of synthetic faces did not effect the output, whereas identities of real faces and lighting of synthetic faces were properly preserved and combined. For the VW-300 domain that contained face crops partially occluded by a microphone, the proposed model preserved size and position of the microphone, and properly manipulated images with partially occluded mouth, even though this attribute was not modeled by the simulation.
    \item Larger discrepancy between attribute distributions in two domains (last two rows in Table \ref{tab:results}) leads to poorer attribute disentanglement, \eg if synthetic digits are much smaller than real (fifth row in Table \ref{tab:results}), or much less size variation is present in the real MNIST (sixth row), or much less rotation in USPS (Figure \ref{fig:supl_usps_rot}). For moderate discrepancies in attribute distributions, AoI (\eg rotation) in generated images followed the distribution of AoI in the real domain (Figure \mref{fig:results}{e}). 
    If during evaluation the property of interest in a synthetic input was beyond values observed during training, model's outputs ``saturated'' (Figure \mref{fig:results}{(iii)}).
    \item Ablation study results (Table \ref{tab:supl_ablation}) and the visual inspection of generated images suggest that domain-agnostic encoders help to semantically align embeddings of attributes across domains. Image level GAN losses improve ``interchangeability'' of embedding components from different domains. Learned representations are highly excessive, so even basic properties such as ``digit rotation'' required double digit embedding sizes. Attribute cycle losses together with pixel-level instance noise in attribute and disentanglement losses improved convergence speed, stability, and the resilience of the model to degenerate solutions.
\end{enumerate}

\textbf{Comparison to Related Methods.} To our knowledge, only E-CDRD \cite{liu2018detach} and DiDA \cite{Cao2018dida} considered similar input constraints at train time (both use explicit labels). We could not obtain any implementation of E-CDRD, and since authors focused on different applications  (domain adaptation for digit classification, manipulation of photos using sketches), their reported results are not comparable with ours. Solutions found by DiDA were degenerate (used only one input) for both rotation and size manipulation of digits (Figure~\mref{fig:munit}{c}), \ie DiDA converged to plain domain translation; the available implementation of DiDA made it very difficult to apply it to faces. While MUNIT \cite{huang2018munit} (unsupervised cross-domain) and Cycle-Consistent VAE \cite{harsh2018disentangling} (single-domain) methods have input constraints incompatible with ours, we investigated how they preforms, respectively, without attribute supervision and in the presence of the domain shift. The Figure~\mref{fig:munit}{a} shows that MUNIT disentangled face orientation as ``content'' and mouth expression as ``style'', as random style vectors appear to mostly influence the mouth. Unfortunately, style embedding spaces of two domains are not semantically aligned, so controlled manipulation of specific attributes (e.g. mouth) across domains is not possible. On digits MUNIT disentangled digit stroke as style and everything else as content (Figure~\mref{fig:munit}{d}), so rotation and size could not be manipulated while preserving the class label. Cycle-Consistent VAE learned great disentangled representations and enabled controlled manipulation of \textit{synthetic} images, but, when applied to images from the real domain, could only encode attributes of real images that resembled synthetic ones (Figure~\mref{fig:munit}{e}), and could not generate plausible real faces because domains looked too different (Figure~\mref{fig:munit}{b}). 



%
\section{Conclusion}
In this paper we present a novel task of ``cross-domain image manipulation by demonstration'' and a model that excels in this task on a variety of realistic and proof-of-concept datasets. Our approach enables controlled manipulation of real images using crude simulations, and therefore can immediately benefit practitioners that already have imprecise models of their problem domains by enabling controlled manipulation of real data using existing imprecise models.




\clearpage

{\small
\bibliographystyle{ieee}
\bibliography{egbib,pgn}
}

\clearpage
\section{Supplementary}

\begin{table}[t]
    \centering
            \begin{tabular}{|c|c|c|c|c|}
\hline
{\small \bfseries Mod} & Acc & $r_{attr}$ & $ r_{rest}$ & $V_{rest}$\\
\hline
{\small original} & 0.94 & 0.4 & 0.11 & 0.01 \\
{\small two enc} & 0.15 & 0.32 & 0.24 & 0.07 \\
{\small one dec} & 0.49 & 0.71 & 0.04 & 0.02 \\
{\small k=16} & 0.14 & 0.8 & 0.5 & 0.25 \\
{\small d=64} & 0.75 & 0.61 & 0.42 & 0.01 \\
\hline
\end{tabular}
\vspace*{16.5px}
    \caption{Ablation study for digit rotation: two encoders instead of single shared encoder, non-shared decoder, smaller rotation embedding (same overall embedding size) and two times reduced dimensionality of both embeddings.\label{tab:supl_ablation}}
\end{table}
\begin{figure*}
    \centering
    
    \supincludegraphics[width=\textwidth]{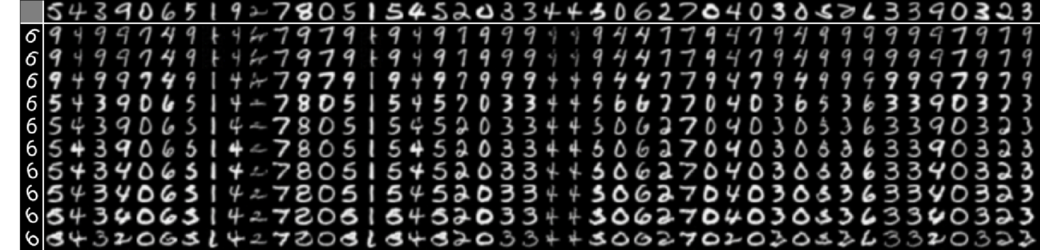}
    \hspace*{20px}\supincludegraphics[width=0.98\textwidth]{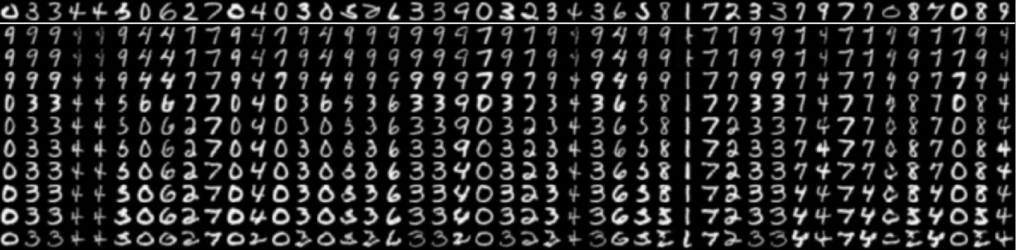}
    \caption{Rotation was not disentangled, probably due to the lack thereof in USPS naturally. 
    \label{fig:supl_usps_rot}}
    
    \supincludegraphics[width=\textwidth]{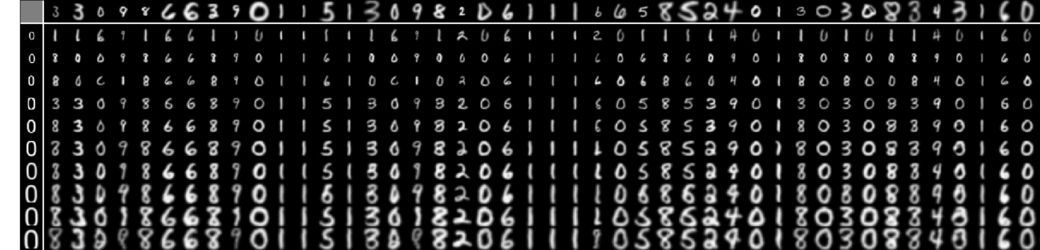}
    \hspace*{20px}\supincludegraphics[width=0.98\textwidth]{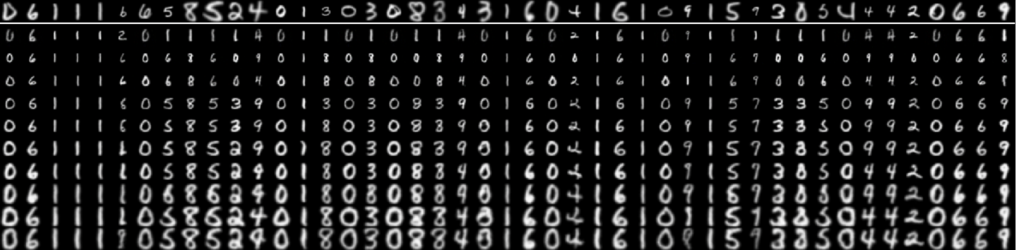}
    \caption{Size disentangled in USPS digit using synthetic renders. \label{fig:supl_usps_size}}
\end{figure*}

Please take a look at our video demonstration available at {\small \url{http://bit.ly/arxiv_puppet}} and in the attached MP4 file (x264 codec). More images with more detailed experiments on all datasets are given below.

\subsection{Implementation details.} 
\textbf{Architecture}. We used the ``CycleGAN resnet'' encoder (padded 7x7 conv followed by two 3x3 conv with stride 2 all with relus), followed by six residual conv blocks (two 3x3 convs with relus) a  fully-connected bottleneck of size 128 and a pix2pix decoder (two bi-linear up-sampling followed by a convolution). We used LS-GAN objective in all GAN losses. It generally follows the architecture of CycleGAN implementation provided in the tfgan package\footnote{ \url{https://www.tensorflow.org/api_docs/python/tf/contrib/gan/CycleGANModel}}. 

\textbf{Training}. We optimized the entire loss jointly with respect to all encoder-decoder weights and then all discriminator losses in two consecutive iterations of the Adam optimizer with $\alpha=\text{(2e-4, 5e-5)}$ learning rates with polynomial decay and $\beta=0.5$. A model trained by updating different losses wrt different weights independently in alternating fashion did not converge, so all generator and discriminator losses must be updated together in two large steps. We also added Gaussian instance noise to each image used in disentanglement and attribute cycle losses to improve stability during training. We added stop gradient op after the application of $C_B$ in the second attribute cycle loss and instance noise to all intermediate images to avoid the ``embedding'' behaviour.

We purposefully avoid constraining embeddings themselves, \eg penalizing Euclidean distances between embedding components of images that are known to share a particular attribute, as such penalties often cause embedding magnitudes to vanishing.

\textbf{Hyperparameters}. The reasonable choice of loss weights we used is given below. We did not perform any large-scale hyperparameter optimization, just tried a couple of combinations, the $\mathcal L_{rest}$ weight required few (2-3) manual tuning attempts to balance $\mathcal L_{attr}$, (\ie tried 1 then 5 then 3). 
\begin{gather*}
    \mathcal L_{total} = 10 \cdot \mathcal L_{rec} + 10 \cdot \mathcal L_{cyc} + 5 \cdot \mathcal L_{attr} + 3 \cdot \mathcal L_{rest} + \\ + \sum_{K_1, K_2, K_3} \sum_{x \in K_1, y \in K_2} \mathcal L^{(K_3)}_{GAN}(C_{K_3}(x, y))
\end{gather*}

\textbf{Metrics}. The quality of attribute isolation can also be evaluated by estimating mutual information between attribute values and parts of embeddings that should or should not encode it \cite{harsh2018disentangling}; we do not explicitly penalize our model for embedding extra information as long as decoder learns to ignore it, so this metric was not useful in our case.

\textbf{Related methods}. In related experiments we used a modified DiDA implementation from \url{https://github.com/yangyanli/DiDA/} in both last and best training modes, MUNIT from \url{https://github.com/NVlabs/MUNIT} and cycle-consistent VAE \url{https://github.com/ananyahjha93/cycle-consistent-vae}.

\subsection{Extended Results}
\textbf{``Saturated'' inputs.} To clarify, by ``model saturates'' we mean that if we pass synthetic inputs with the AoI value beyond what we used during training, model outputs reasonable ``highest'' or ``lowest'' output for respective domains instead of breaking (it could, since inputs are not typical).

\textbf{Digits.} You can find results for USPS in Figures \ref{fig:supl_usps_rot} and \ref{fig:supl_usps_size}, model did not manage to disentangle rotation in USPS probably due to the lack of thereof.

\textbf{300-VW.} In addition to the attached video, static examples of manipulated faces can be found in Figures \ref{fig:supl_faces_1}, \ref{fig:supl_faces_2}, \ref{fig:supl_faces_3} and \ref{fig:supl_faces_4}. As pointed in the main paper, model properly preserves orientation and expression of the real input, and mouth expression of the synthetic input, and completely discards everything else.

\textbf{YaleB.} One can find more examples in Figures \ref{fig:supl_yale_1}, \ref{fig:supl_yale_2}, \ref{fig:supl_yale_3} and \ref{fig:supl_yale_4}. Identities of real inputs are preserved most of the time (as a reminder, all YaleB images were combined into a large single domain with no identity labels, so the model had to learn to disentangle and preserve identity). In cases when too little light is available in the real scene, the model ``hallusinates'' an ``average'' identity details. When model is asked to ``imagine'' lighting conditions that were not present in YaleB, but present in the synthetic dataset, some identity details are corrupted.

\textbf{Synthetic faces.}
In Figure \ref{fig:supl_syn_light} we present more examples of outputs of a model trained to disentangle lighting across synthetic identities. ``Dot artifacts'' disappear if model is trained long enough.

\textbf{Color blind and print friendly.} An alternative version of Figure 4 is given in Figure \ref{fig:model_colorblind}.

\newlength{\vwwidth}
\setlength{\vwwidth}{0.48\textwidth}

\begin{figure*}
    \centering
    \supincludegraphics[width=\vwwidth]{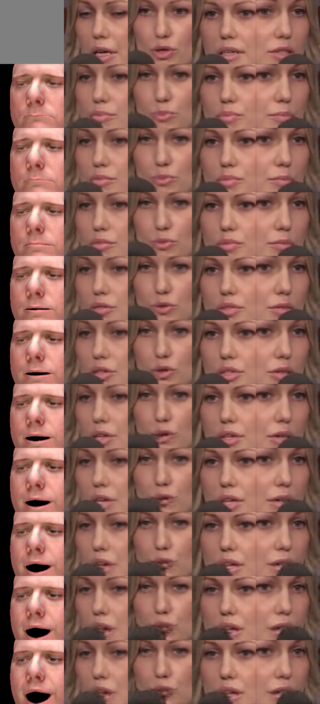}
    \supincludegraphics[width=\vwwidth]{{"imgs/face_supl/flat_summaries__1"}.png}
    \caption{More random examples for an identity from 300-VW dataset with mouth expression manipulated using our model. Two first and two last rows are ``saturated'' examples.
    \label{fig:supl_faces_1}}
    
\end{figure*}

\begin{figure*}
    \centering
    \supincludegraphics[width=\vwwidth]{{"imgs/face_supl/flat_summaries__3"}.png}
    \supincludegraphics[width=\vwwidth]{{"imgs/face_supl/flat_summaries__4"}.png}
    \caption{More random examples for an identity from the 300-VW dataset with mouth expression manipulated using our model. Two first and two last rows are ``saturated'' examples.
    \label{fig:supl_faces_2}}
    
\end{figure*}

\begin{figure*}
    \centering
    \supincludegraphics[width=\vwwidth]{{"imgs/face_supl/flat_summaries__6"}.png}
    \supincludegraphics[width=\vwwidth]{{"imgs/face_supl/flat_summaries__7"}.png}
    \caption{More random examples for an identity from the 300-VW dataset with mouth expression manipulated using our model. Two first and two last rows are ``saturated'' examples.
    \label{fig:supl_faces_3}}
    
\end{figure*}

\begin{figure*}
    \centering
    \supincludegraphics[width=\vwwidth]{{"imgs/face_supl/flat_summaries__9"}.png}
    \supincludegraphics[width=\vwwidth]{{"imgs/face_supl/flat_summaries__10"}.png}
    \caption{More random examples for an identity from the 300-VW dataset with mouth expression manipulated using our model. Two first and two last rows are ``saturated'' examples.
    \label{fig:supl_faces_4}}
    
\end{figure*}

\begin{figure*}
    \centering
    \supincludegraphics[width=\textwidth]{{"imgs/yale_supl/flat_summaries_2"}.png}
    \caption{More random examples for a single identity from the YaleB dataset with lighting expression manipulated using our model.
    \label{fig:supl_yale_1}}
    
\end{figure*}

\begin{figure*}
    \centering
    \supincludegraphics[width=\textwidth]{{"imgs/yale_supl/flat_summaries_4"}.png}
    \caption{More random examples for a single identity from the YaleB dataset with lighting expression manipulated using our model.
    \label{fig:supl_yale_2}}
    
\end{figure*}

\begin{figure*}
    \centering
    \supincludegraphics[width=\textwidth]{{"imgs/yale_supl/flat_summaries_14"}.png}
    \caption{More random examples for a single identity from the YaleB dataset with lighting expression manipulated using our model.
    \label{fig:supl_yale_3}}
    
\end{figure*}

\begin{figure*}
    \centering
    \supincludegraphics[width=\textwidth]{{"imgs/yale_supl/flat_summaries_10"}.png}
    \caption{More random examples for a single identity from the YaleB dataset with lighting expression manipulated using our model.
    \label{fig:supl_yale_4}}
    
\end{figure*}

\begin{figure*}
    \centering
    \supincludegraphics[width=\textwidth]{{"imgs/yale_supl/flat_summaries_8"}.png}
    \caption{More random examples for a single identity from the YaleB dataset with lighting expression manipulated using our model.
    \label{fig:supl_yale_4}}
    
\end{figure*}

\begin{figure*}
    \centering
    \supincludegraphics[width=\textwidth]{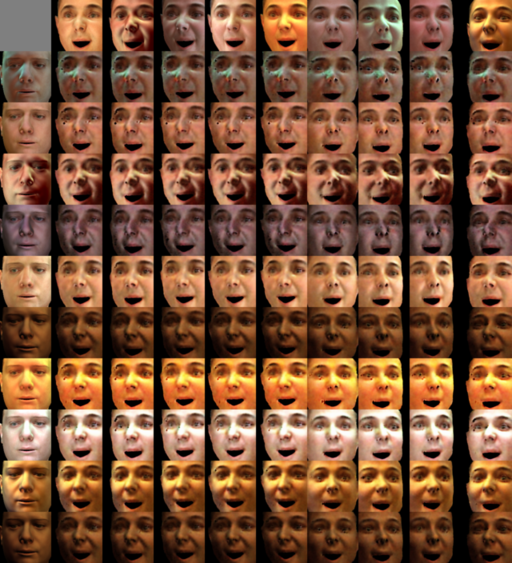}
    \caption{More random examples for disentanglement of spherical harmonics across synthetic identities.
    \label{fig:supl_syn_light}}
    
\end{figure*}
\begin{figure*}[ht]
\centering
\vspace*{-5px}
\supincludegraphics[width=\textwidth]{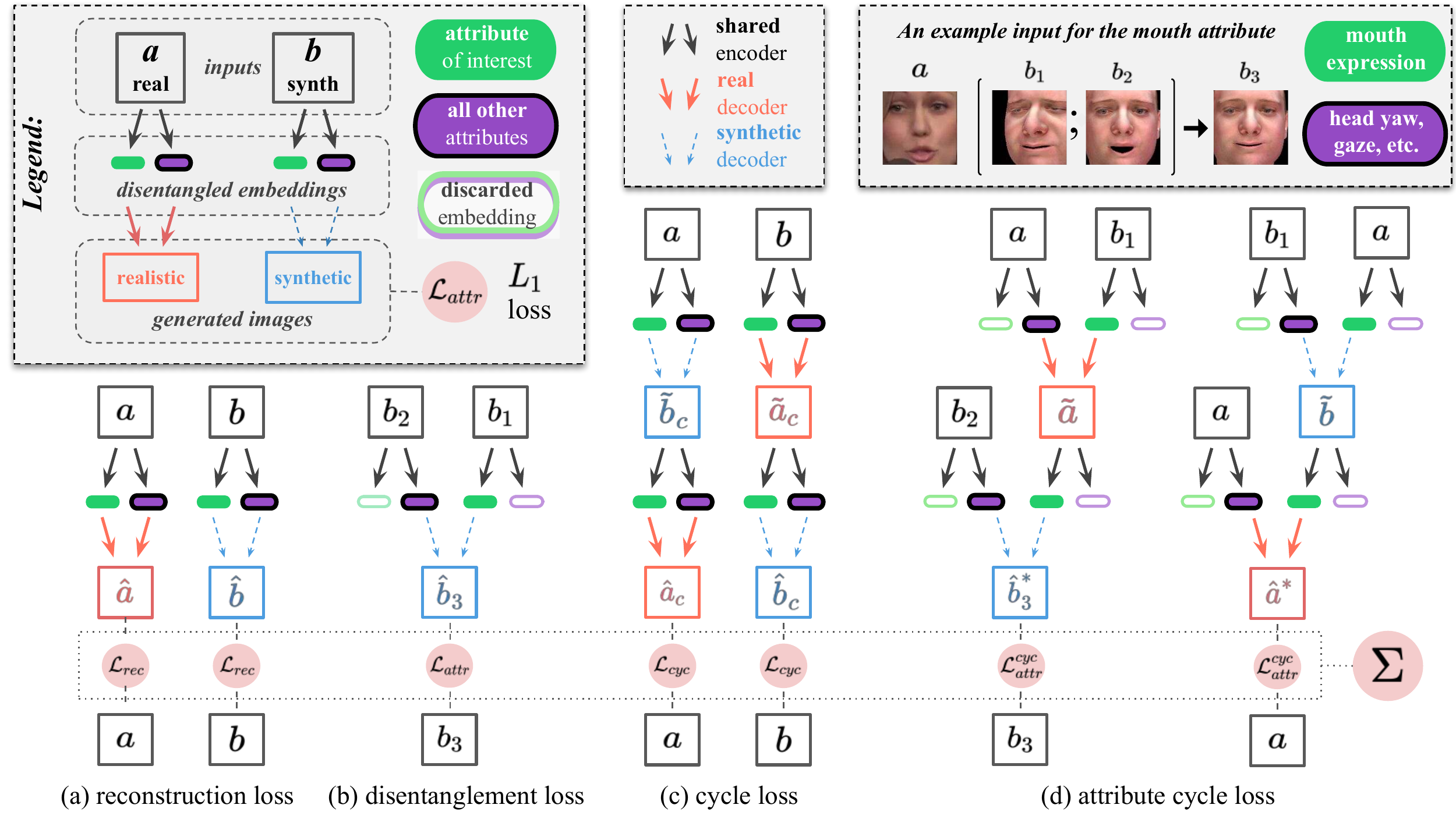}

\caption{\textit{(A color blind and print friendly version).} Supervised losses jointly optimized during training of the PuppetGAN. When combined, these losses ensure that the ``attribute embedding'' ({\color{ForestGreen} green} capsule without a border) affects only the attribute of interest (AoI) in generated images, and that the ``rest embedding'' ({\color{Plum}purple} capsule with a bold border) does not affect the AoI in generated images. When trained, manipulation of AoI in real images can be performed by replacing their attribute embedding components. Unsupervised (GAN) losses are not shown in this picture. An example at the top right corner illustrates sample images fed into the network to disentangle mouth expression (AoI) from other face attributes in real faces. Section 3 provides more details on the intuition behind of these losses. \vspace*{-10px}}
\label{fig:model_colorblind}
\end{figure*}

\end{document}